\documentclass[runningheads]{llncs}
\usepackage{amsmath,amssymb,amsfonts}

\usepackage{algorithmic}
\usepackage{graphicx}
\usepackage[ruled, vlined]{algorithm2e}

\usepackage{subfigure}
\usepackage{textcomp}
\usepackage{xcolor}
\usepackage{url}
\usepackage{cite}
\DeclareUnicodeCharacter{2212}{-}

\begin{document}
\title{Conversational Pattern Mining using Motif Detection}
%
%
\author{Anonymous}
\author{Nicolle Garber \and
Vukosi Marivate\orcidID{0000-0002-6731-6267}}

\institute{Department of Computer Science\\ University of Pretoria\\ 
\email{NikiGarber@gmail.com},
\email{vukosi.marivate@cs.ac.za}
}

\maketitle              
\begin{abstract}
The subject of conversational mining has become of great interest recently due to the explosion of social and other online media. Supplementing this explosion of text is the advancement in pre-trained language models which have helped us to leverage these sources of information. An interesting domain to analyse is conversations in terms of complexity and value. Complexity arises due to the fact that a conversation can be asynchronous and can involve multiple parties. It is also computationally intensive to process. We use unsupervised methods in our work in order to develop a conversational pattern mining technique which does not require time consuming, knowledge demanding and resource intensive labelling exercises. The task of identifying repeating patterns in sequences is well researched in the Bioinformatics field. In our work, we adapt this to the field of Natural Language Processing and make several extensions to a motif detection algorithm. In order to demonstrate the application of the algorithm on a dynamic, real world data set; we extract motifs from an open-source film script data source. We run an exploratory investigation into the types of motifs we are able to mine.

\end{abstract}
\section{Introduction}
Motif discovery can be phrased as the detection of `functionally significant short, statistically overrepresented subsequence patterns in a set of biological sequences’ ~\cite{kohler2012improved}. More simply, we can think of a motif as an abstract generalisation of short repeating subsequences in a collection of sequences. These repeating subsequences contain variations from each other and from the motif that tries to describe them. Technically, this introduces challenges. However; when applied to conversations, this allows us to capture various dynamics and different ways in which people's conversation can evolve. The focus of our work is to capture repeating conversational patterns through the use of motif detection. We demonstrate this by focusing on open source film scripts of a comedy genre. 

Successful motif detection in conversations can prove useful in many applications. Critically, finding conversational cues can assist with intelligent conversation labelling. This has wide applications in various business domains with service centres dealing with a customer base through text using chat-bots, customer agents or even phone transcripts. We can start to think of instructive cues for chat-bots, determining themes of customer sentiment development or service requests. On social media platforms such as Twitter, mining motifs may uncover recognizsable cues which might assist in event detection. In any context, quantifying conversation gives us the power to abstract, index and search. This helps us to extract important things in any domain.

We turn to the field of Bioinformatics, in which motif detection forms a large and fundamental part of understanding important biological sequence structures such as DNA and proteins. We attempt to transfer the idea of a motif representing a structural unit into the idea of a structural unit of a conversation. 

Conversation in a real world goal-oriented setting (such as in business), contains a particular set of goals and therefore is likely to have a more limited set of conversational cues. It is therefore expected that clearer motif patterns might emerge. Likely, if it is observed that conversational motif mining is successful on a more varied dataset, we can have increased confidence in its application in a more targeted setting. We simulate this idea by application in a broad and varied conversational setting such as an open-sourced dataset of film scripts. We focus our attention on one particular genre of the film scripts in order to constrain the diversity of the underlying conversations. It is well known that the comedy movie genre makes good use of tropes and movie motifs \cite{mcdonald2007romantic}. 

In addition, it is clear after some experimentation that the comedy genre has the largest set of conversations of any genre, which allows for richer sequence data in order to run motif detection. We investigate the types of motifs we are able to extract.

There are two larger components that constitute the motif mining pipeline. Firstly, we need to convert the conversational text data into meaningful sequences. The second part of the pipeline consists of the application of motif detection on top of these sequences. Our paper is structured as follows: the second section provides a view of previous work done in the Bioinformatics and Natural Language Processing (NLP) domains. Next, we describe the motif mining pipeline in the methodology section. The results section is split into two parts: the first deals with the conversion of conversations into sequences while the second deals with the motif mining algorithm. We show the reader a view of the motif mining algorithm on generated data after which we investigate the film scripts dataset results. Finally, we follow a brief summary, discussion of the method, results and possible extensions in the conclusion. 

\section{Background}
In this section, we cover the concept of a motif, how it has been used and how we would like to leverage it. The reader might be interested to note that while motif detection associates very highly with the field of Bioinformatics, the concept is being applied in a number of different areas. For example, motifs can help to uncover network structures which can find application not only in Bioinformatics but in communications and software engineering \cite{meira2018improved}. The former two are active areas of research and there are several survey papers on both of these \cite{ciriello2008review}, \cite{ribeiro2009strategies}, \cite{wong2012biological} and \cite{das2007survey}, \cite{hu2005limitations} for example. Motif detection has been used in \cite{kirschbaum2018lemonade} by the authors to understand information processing in the brain via motif detection through variational autoencoders. Motif learning was used in a very interesting way by the authors of \cite{zhao2019motif} who use motif-based methods in order to discover higher-order similarities in heterogeneous information networks for use in recommender systems.

A motif is a characterization of a short subsequence of DNA \cite{das2007survey}. This sequence typically indicates a significant biological structure such as DNA binding sites for regulatory proteins \cite{das2007survey} (a common application in the field).
Segments of DNA which are used for transcription processes are called genes and the information contained in these helps with making proteins. In many cases, the transcription process is a function of the genes of a sequence and finding these patterns is one of the most important and challenging fields of molecular biology and computer science \cite{das2007survey}.

The challenges of motif detection lie in the fact that any real data as in DNA sequence data (or in our case conversational sequences) contains noise. In biological sequences, this is expressed as mutations, insertions or deletions of nucleotides \cite{das2007survey}. Additionally, motifs can either occur in the same gene or in multiple genes. Some genes may not have them \cite{das2007survey}. More traditional techniques search for exactly one motif per sequence. This may miss some motifs or conversely detect motifs in sequences which may not contain them. Since most motif detection algorithms look for some type of global commonality between motifs in sequences, sequences that do not have this motif would add noise to this process. For these reasons, motif detection algorithms typically need to be more complex than pattern matching or brute force algorithms. 

We focus on a category of earlier approaches used which were found to work better with longer motifs \cite{das2007survey}. These are probabilistic models which make use of some type of position weight matrix. Each position of a motif is represented by the various probabilities of a `letter' (sequence element) in that position. This type of structure deals better with weakly constrained positions where variation is likely \cite{das2007survey}. These methods are not guaranteed to find a global optimum and make use of more local search technique methods \cite{das2007survey}. An example of a way to measure a subsequence as a motif has been described by Tompa as explained by the review paper authors of \cite{das2007survey}. The probability of finding a motif $s$ in $N$ number of sequences can be characterised with a statistical significance test which includes the formulation of a z-score. This is asymptotically normally distributed with mean 0 and standard deviation 1. This score makes it possible to compare different motifs. The z-score is as follows: \begin{equation}
M_s=\frac{(N_s - {NP}_{s} )}{\sqrt{(Np_s )(1 - p_s ) }}
\end{equation}
where $N_s$  is the number of sequences containing occurrence of $s$ (sequence motif) and $p_s$  is the probability of at least one motif $s$ in a sequence \cite{das2007survey}. While there have been more sophisticated motif detection methods developed, our work focuses on some of the earlier or more classical techniques. This is intended as an introductory experiment into the concept of motif mining in conversations. 
 
\subsection{Conversational Mining}
Analysing conversations can be thought of in the context of the broader field of Natural Language Processing (NLP). There is no single accepted definition of NLP; however, the general meaning is quite consistent. That is: a set of computational techniques aimed at understanding, summarising and representing text. This is done with the aim of striving towards human-like capabilities of generating and processing text for a diverse set of tasks \cite{6786458}.

The most common type of work done in computational fields with regards to conversations is various methods and enhancements for dialogue systems which maintain dialogue state and can be used for generative conversational agents. The survey paper \cite{chen2017survey} performs a good investigation of advances in the space by dividing dialogue systems into task-oriented and non-task oriented models. The work in \cite{gavsic2017spoken} focuses on surveying different aspects of spoken language understanding in order to build better conversational systems. The recent work \cite{szpektor2020dynamic} addresses building a domain driven bot by separating content selection for candidate responses from response composition. This was integrated into the Google assistant. The task of dialogue generation is approached in \cite{chen2018hierarchical} by a hierarchical variational memory network. This aims to abstract variations as well as long-term dependency modeling such as is necessary for dialogue state tracking. A different approach is followed in \cite{li2019dense} where multi-turn conversation is modeled using a dense semantic matching network which leverages context-response pairs to find an optimal candidate. An important aspect of research in dialogue systems includes examining evaluation metrics of generative models. The authors of \cite{liu2016evaluate} provide recommendations for building metrics that perform better than the technical metrics created for machine translation in this domain.  

Some earlier work \cite{lipizzi2015extracting} places focus on performing analysis on conversational patterns in social media. The focus of the work is the examination of product launches on Twitter. Conversational analysis is leveraged in order to create concept maps to compare conversations. A concept map is defined as a network that follows the conceptual progression of a conversation (in this context) by keyword representation which can lead to the discovery of topic clusters which are related to each other. The approach appears to be quite handcrafted - requiring specific pre-processing for the domain and the creation of domain-specific dictionaries. Traditional network analysis is applied in order to extract the concept map and topic flows. Semantic complexity cannot be modelled through single keywords and the abstract structures that are extracted are vague. The notion of conversational structure abstractions is not brought forward. For example, the abstraction is extracted in the form of a word cloud. Another example of analysis-style work is from a more recent paper which leverages a social media platform Gab in order to detect `echo-chamber' patterns by \cite{bagavathi2019examining}. The structural abstraction of conversations which we extract in the form of motifs is extracted in this work in the form of cascades. This is a directed graph where the set of vertices represent content (posts, replies, quotes) and the edges represent replies and quotes (or interactions with content). This structure evolves over time with the original post forming a `root node' of sorts and subsequent interactions form levels. 
While these cascades are able to model linear and non-linear interactions, there is only brief work done on modelling the content and conversational queues (in the form of hashtags) and most emphasis is placed on response time and spread modelling of the different types of cascades.

We see that motifs are not applied to conversations, to the best of our knowledge. In fact, the conversational pattern mining field appears saturated with work on dialogue management systems, bots, assistants, dialogue act prediction and question answering. There seems to be an opportunity for creating more data mining techniques for conversation. We can do this in order to begin understanding generic structures of conversations for modelling or extraction. It seems as though there is less work developed on converting conversations into either embeddings (as we have done) or even discrete representations over time. Motifs assist with understanding generic conversational structures which can be used for a host of use cases - from segmentation of important parts of conversation, to search, to understanding, to prediction, to visualisation to clustering. With the understanding of generic structures of conversation, just as in Bioinformatics, we can begin to assess individual deviation from global structures, build anomaly detection, understand human conversations in a domain better - particularly conversational queues in goal-oriented domains or even queues displaying negative conversational patterns to detect on social media.

\section{Methodology}
\begin{figure*}[t]
  \includegraphics[width=\textwidth]{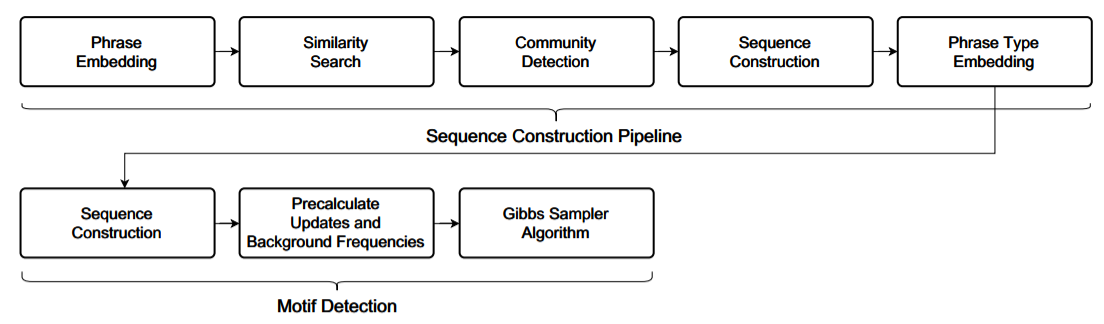}
  \caption{Modular Representation of Pipeline for Conversational Motif Mining Strategy}
  \label{fig:Pipeline}
\end{figure*}

We begin our pipeline by structuring conversational text into sequence data. We present the higher level pipeline components in Figure~\ref{fig:Pipeline}.

We have two larger pipeline components: sequence creation and motif detection. In the sequence creation component, we find embeddings for the phrases on a sentence level. In the traditional use of motif detection, the sequences consist of categorical elements. We aim to create an analogue to this by grouping similar phrases together to form generic phrase types. The volume of data produced during this process necessitates an efficient clustering technique. We therefore perform community detection on phrases and their most similar neighbours. The communities found in this way are representative of similar phrases. For simplification, we produce a low dimensional representation of each phrase type by finding the centroid of the classes and performing dimensionality reduction. This forms the sequence base upon which we are able to run the motif detection pipeline. 

The motif detection pipeline is built on the Gibbs sampler framework \cite{lawrence1993detecting}. 
We describe the extensions needed for transferability from the Bioinformatics to the conversational text domain. 

\subsection{Sequence Creation Process}
From each of the spoken phrases, we aim to find a generic ``phrase type". This is: an abstract class capturing the semantic, emotional and conversational meaning of a phrase within the context of the conversation. 
The high-level steps in this process look as follows:

\begin{enumerate}
    \item Phrase level embeddings are constructed.
    \item A graph is created using the embeddings. Communities within this graph are extracted and represented as centroids.
    \item Simpler representations are created using dimensionality reduction.
\end{enumerate}

\subsubsection{Embedding Construction}
We find phrase level embeddings in order to capture the dynamics of spoken phrases in a conversation. This is done by performing word level embeddings, followed by sentence embeddings. The good structure of the text we are dealing with (minimal spelling mistakes, special characters and mixed languages) enables us to leverage pre-trained embeddings. This avoids the resource expensive task of training our own embeddings or fine-tuning. 

Word vector representations are derived using the well known continuous bag of words (CBOW) model where each word is represented by a bag of character n-grams \cite{mikolov2018advances}. From here, phrase embeddings are obtained by encoding the FastText word embeddings with the InferSent model \cite{conneau-EtAl:2017:EMNLP2017}. 

\subsubsection{Community Detection for Creation of Phrase Classes} For the creation of the classes representing `phrase types', we turn to unsupervised methods. 

Each of the spoken utterances is represented by a high dimensional vector $x \in \mathbb{R}^{4096}$. 
This vector contains a lot of information (4096 dimensional space). The motif detection algorithm works by scanning over multiple sequences with a moving window and finds optimal alignment by looking at where the motifs achieve the greatest similarity. Working with four thousand dimensional sequences is infeasible for analysis and it is not interpretable. Each sentence embedding on its own does not tell us the abstract conversational class this phrase belongs to. We have a continuous spectrum of vectors $x \in \mathbb{R}^{4096}$ which do not capture a more hierarchical structure. 
We construct abstract classes with information about the conversational function of the class, for example: greeting phrases, displays of empathy, pause cues. In order to find these classes, we perform semantic grouping. Since the dataset is too large to perform simple clustering on, we perform the following succession of tasks: 
\begin{enumerate}
    \item Run an index on the dataset for fast KNN query \cite{DBLP:conf/sisap/BoytsovN13} (we use NMSLIB with angular distance).
    \item For each phrase in the dataset, search for the 10 nearest neighbours based on angular distance.
    \item Create a dataset in which each row has phrase, neighbour and angular distance. This forms the structure on top of which a graph is built. The phrase is the source node, the neighbour is the target node and the edge weight is represented by angular distance. 
    \item Extract communities from groups of common phrases that form. The community is represented as a centroid of the embeddings of the phrases within it. This is done by making use of the community info\_map algorithm \cite{Rosvall_2008}.
\end{enumerate}

After obtaining phrase embeddings which represent the nodes in our graph, we find the edges by detecting the 10 closest neighbours by angular distance. The KNN classifier as described by \cite{5408784} is a decision rule based classifier which assigns a classification to an unlabeled point based on the classification of the nearest classified points. We make use of NMSLIB \cite{DBLP:conf/sisap/BoytsovN13} : Non-Metric Space Library which is specifically designed for efficient similarity search based on various non-metric spaces. It is very fast and efficient. We make use of the angular distance defined by:\[d(x,y)= \arccos{(\frac{\Sigma_{i=1}^{n}x_iy_i}{\sqrt{\Sigma_{i=1}^{n}x_i^2}\sqrt{\Sigma_{i=1}^{n}y_i^2}})}\] It is  found that this distance works the best when dealing with our embeddings. Additionally, it is known that cosine distance works well with embeddings produced by models where the objective function was represented by a dot product with softmax. The algorithm uses Hierarchical Navigable Small World Graphs (HNSW) which builds a representation of stored elements incrementally through proximity graphs. This search technique has a logarithmic complexity with very high performance. The KNN algorithm is implemented with search for the 10 nearest neighbours of every phrase.

By utilising the KNN techniques, we find close relationships of phrases. If many phrases share many neighbours this will be well captured in a graph structure. Phrases fall into communities where there are many similar phrases. Because we model this on a phrase level, these will capture conversational queues. In addition, because of the quality of embeddings used, any semantic similarity, as well as syntactic similarity is captured within the communities.

The community info\_map algorithm was first introduced in 2008 by the authors of \cite{Rosvall_2008}. The idea here is to minimise the map equation (objective function) over all partitions of the network. We use this algorithm as a base to obtain our community clustering (which in our application case consists of 3232 communities). There are many smaller communities which may be very similar to larger ones.
We use the extracted centroids of the communities with dimensionality reduction as opposed to using categorical representations for the communities for the following reasons: 
\begin{enumerate}
    \item We lose relational information of clusters relative to other clusters.
    \item We lose meaning about the clusters.
    \item We are forced to deal with the clusters in a categorical manner which has many implications for the motif detection algorithm.
\end{enumerate}

\subsubsection{Simplification of Phrase Class Representation} We have calculated `phrase classes' in which each phrase can map to a more abstract representation. As we have discussed above, we map each class to a vector by calculating the centroid of each phrase within a community. The challenge at this point is to find a dimensionality reduction technique that will adequately represent each community in relation to all other communities. We focus on conserving the pairwise distances between the centroids. Manifold learning is a good approach for this task as we are dealing with a high dimensional space which will benefit from a non linear method. In order to select the best algorithm, we select the one that maximises the correlation of pairwise cosine distances before and after reduction. We tune the parameters by grid search. The Uniform Manifold Approximation and Projection (UMAP) algorithm developed in \cite{lel2018umap} is a non-linear dimensionality reduction technique. It enjoys good speeds (scalability) and a greater preservation of global structure. We now have representations of phrases in conversational sequences. We map each phrase into its `phrase type' class. This class has a vector representation in five dimensions in a euclidean space with cosine distances preserved.

\subsection{Motif Detection}
In this part of the pipeline we focus on detecting the development of conversational cues that follow a generic pattern common to most or all of the movies.

We look at the Gibbs sampling method \cite{lawrence1993detecting}. Gibbs sampling methods have been used at length for motif detection. The authors describe the benefits of the implementation as $N$ linear run time for $N$ sequences as well as allowing for increased variability among the patterns that are found. The latter is useful for modelling human conversations which are not a prescribed set of exact rules. For example, in a standard greeting sequence, there may be a generic pattern which includes the following turns: 
\begin{enumerate}
    \item Character1: greeting phrase
    \item Character2: greeting phrase 
    \item Character1: generic response phrase
    \item Character2: generic response phrase
    \item Character1: queue for initiation of specific topic
    \item Character1: response of initiation of specific topic
\end{enumerate}
If a third character is introduced, the pattern will likely change. However; it will still form an instance of a generic greeting queue. In gene sequences, this variation is caused by factors such as mutation and as such it is a consideration of the underlying algorithm. 

\subsubsection{Gibbs Sampler Algorithm}
In the base algorithm \cite{lawrence1993detecting}, each sequence is assumed to have exactly one motif. The Markov assumption is also applied. Given a set of $N$ sequences ${S_1,...,S_N}$, we look for segments of the sequences which are most similar to each other. The segment is of fixed length $W$. The measure of similarity of the sequences is one which maximises the ratio of pattern (motif) probability to background probability. When we have a good alignment of motifs or positions in each sequence where the segment maximises this ratio, we stop the iterations of the algorithm.

This ratio is: \[A_x=\frac{Q_x}{P_x}\] 
$A_x$ is described as the probability that a sequence $z$ contains a motif at position $x$ (foreground probability over background probability). At each iteration, one of the sequences is held out. The window is tested at each position of the sequence by calculating the probability $P_x$ of generating the sequence using background probability. The probability profile elements $q_{i,j}$ \cite{lawrence1993detecting} are described by: 
\begin{equation}
q_{i,j} = \frac{(c_{i,j} +b_j)}{(N - 1 + B)}
\end{equation}  where the $c$ term represents the count of amino acids that occur in the $i^{\text{th}}$ position and the $b$ term is the background frequency of the amino acid. A random segment is selected from all the window candidates according to the weights $A_x$ at each $x$. Over time, $q_{i,j}$ begin to reflect a pattern existing in other sequences. The segment probabilities $A_x$ become more strongly characterised and the algorithm tends to favour further positions in other sequences that confirm these patterns.

There are several constraints of the motif algorithm. These include:
\begin{itemize}
    \item Constraining each sequence to have exactly one instance of a motif.
    \item A fixed pattern of length $W$.
    \item A fixed sequence length.
    \item The algorithm is stopped after a number of iterations as opposed to a convergence criterion. 
\end{itemize}

These are considerable limitations which have practical effects not just on our particular application but on most applications we could consider motif mining for. In particular for real life scenarios, we may have some sequences like some time series or conversations which we think may contain a pattern. We don't know beforehand what the pattern looks like or which sequences have it. The first constraint means that sequences that do not have the same pattern as others get considered equally to others and may throw off the pattern detection because it would affect global alignment scores. It also means that if a sequence has more than one instance of the same pattern, this would not be picked up by the algorithm - a separate match algorithm would have to be run post hoc. A fixed motif size is difficult to fine tune. We may not know what type of motif we are looking for and there may be a vast range of possibilities to choose from. Fixed motif lengths may also impose too tight a constraint on the detection of a global pattern against local sequence patterns. For example, some sequences may express a motif with more noise in between its elements and be longer than others. The next constraint, a fixed sequence length is at times not a practical limitation. In our application, our movie conversations are of different lengths. In order to shape sequences to the same size we may need to add extra consideration to how and where we cut them and this might result in information loss. The last assumption is a beneficial one which allows us to consider using the method in the first place. There is support for variation within the motifs which is ideal for conversations and any time series, as we have mentioned before. The fact that the algorithm is heuristic but does not have built in convergence criteria imply that it is fairly easy to stop the algorithm prematurely or conversely have very long and unnecessary run times which add marginal or no improvement. 

We build on the implementation of \cite{sczopek} with some extensions. Firstly, we address the challenge of implementing motif detection in a completely different setting such as conversational mining. Secondly, we make extensions to the original algorithm that allow it to work not only in this setting but in other multivariate time series settings. 
This may allow a practitioner to overcome a seasonality constraint in searching for time series patterns and look for irregular patterns in a way that is more robust to noise. We start the description with a high level explanation of the additions we have made and then a brief holistic discussion of the entirety of the algorithm.
We make the following adjustments:
\begin{enumerate}
    \item Adaptation for use on any sequences (In Bioinformatics there is a finite set of elements in the sequences (typically ``A", ``T", ``G", ``C").
    \item Adaptation for use on varied sequence lengths.
    \item Adaptation for algorithm to consider sequence elements which are not categorical.
    \item Extraction of a global motif which characterises most of the underlying instances.
    \item Ability to calculate a local sequence alignment to the global pattern.  
\end{enumerate}

A high level description of the algorithm is as follows: we randomly hold out one sequence at a time and find an optimal window placement on  the sequence (what we refer to as local alignment) while determining its impact globally (global alignment). This is done iteratively and randomly, allowing window selection on each hold out sequence to start improving global alignment and thereby improve local alignment and so on. The probabilistic implementation allows for variations in local alignment while still finding optimal global alignment. The set of best motifs are only updated when a motif on the holdout sequence is found to improve the global score. The algorithm begins with random initialisation of a set of windows on each sequence. These windows are of length $k$ - the motif length. We then take our set of sequences and select a random hold out sequence. We calculate an optimal placement on the sequence by sliding the window on the sequence and choosing a best probabilistic score. Once this window is selected on the holdout sequence, its impact in relation to all the other windows on all the other sequences is assessed. If this global score is improved by the hold out sequence contribution, the motif for the hold out sequence is replaced, otherwise the procedure continues and another holdout sequence is selected. This is following the Gibbs sampling procedure. 

\subsubsection{Extensions}

\begin{figure}[ht!]
\begin{center}
\includegraphics[width=1\columnwidth]{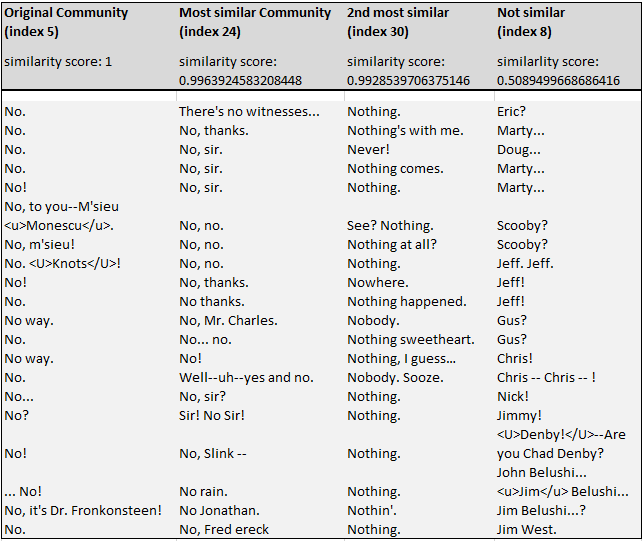}
\end{center}
\caption{Illustration of the need to consider the closeness of our conversational classes}
\label{Community_Similarity}
\end{figure}

In transitioning from the case of DNA bases or amino acids which typically involve 4 classes to a setting in which we have thousands of classes, we consider some extensions. We adjust for the class probabilities tending to zero as the ratio of classes to sequence length grows. The probability of observing any element is obtained from an updated dictionary. Any observation of the element itself adds a count of one. The most similar elements to it (elements with a cosine similarity of greater than cut-off of 0.995) add a proportionately scaled partial count of less than one. When constructing the probability profile for each position, we add all observed instances of the element and all similar elements across the sequences in this position. We also no longer consider a joint probability of observing all the elements of the current window in the motif set. Instead, we calculate the marginal probability of these elements. These probabilities are weighted with an update scheme. For every element in the vocabulary, we add the probabilities of all the most similar elements as observed by a cosine similarity greater than a selected cut-off point. This updated dictionary is precomputed. An example of the updates can be seen in Figure~\ref{Community_Similarity} 

We calculate the global pattern which is used for comparison of each motif in the current set. Each position considers the element with the heaviest count (similarity updates considered). Each position is compared across motifs to the global pattern. We transform it to scale between zero and one. We construct a score vector which accumulates these alignments per position. Each position has the added similarities and is normalised by dividing by the number of motifs. The score vector is then transformed to a single global score by summing individual elements and dividing by $k$. The score vector and $k$ are interpretable. The global pattern is instructive in the understanding of the dataset and the patterns that are found. This structure tells us what motif abstraction looks like. We are also able to cluster our sequences based on how well each motif aligns locally to the global pattern and use this to select the highest aligning local patterns. 


\section{Results}

In this section, we consider an experiment which demonstrates the functionality of the motif detection algorithm. Subsequently, we look at the performance of the conversational motif detection pipeline on a real world dataset. 

\subsection{Motif Detection Results}
\begin{figure}%
\subfigure[Sample of two dimensional Sequences]{%
\label{fig:test3_sequences}%
\includegraphics[width=\columnwidth]{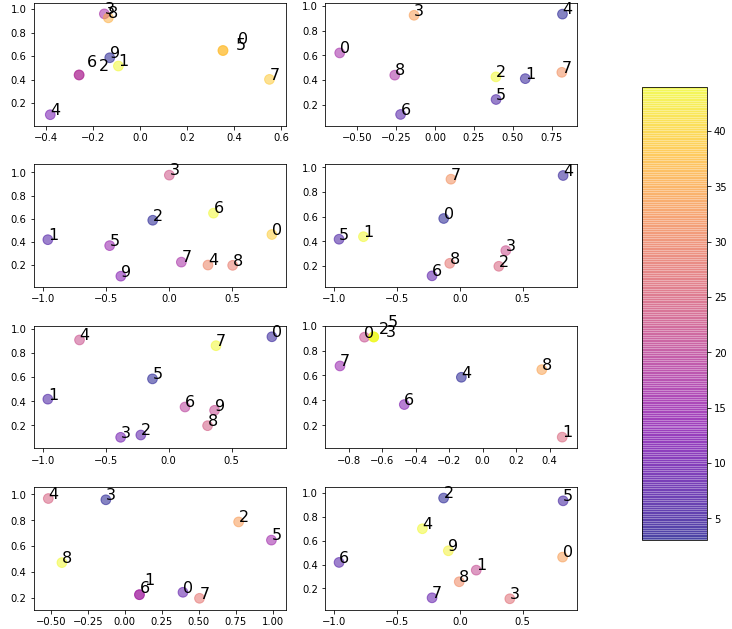}}%
\qquad
\subfigure[Visual results of motif pattern detected]{%
\label{fig:test3_sequences_results}%
\includegraphics[width=\columnwidth]{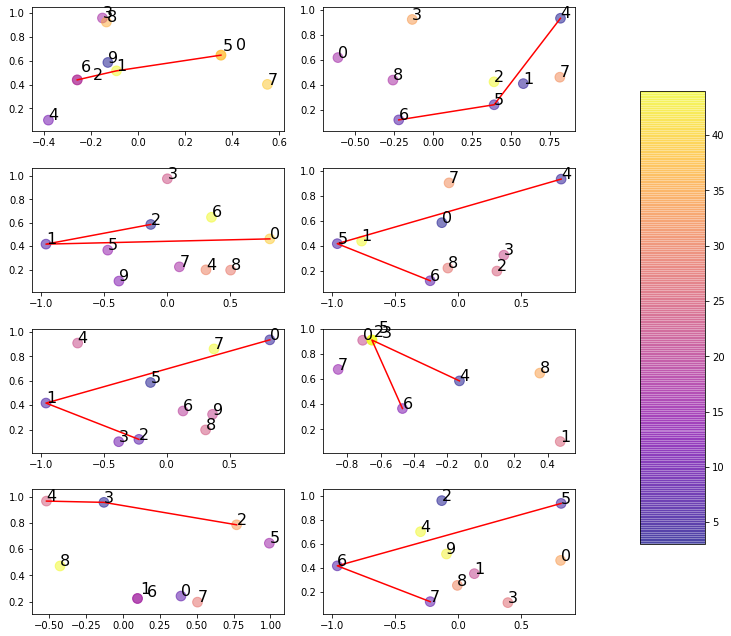}}%
\caption{sample}
\end{figure}

In order to demonstrate how motif detection works, we generate data with motifs and observe whether they are detected. The dataset generated mimics what the motif detection module would expect in our pipeline. We generate 22 sequences and introduce 50 vocabulary' elements. Each element is a uniformly random two dimensional vector: $x \in [-1,1], y \in [0,1]$. The motif elements are generated not to be too similar to each other by creating vectors no more than 0.4 cosine similarity. The representative classes of the motif elements are [3, 5, 7]. With this set of sequences, we run the motif detection algorithm. We visually display the results in Figure~\ref{fig:test3_sequences}:


We detect every instance of an artificially planted motif except one. The global pattern score is 0.88993719 and the global pattern is as expected: 3,5,7. We can see in Figure~\ref{fig:test3_sequences_results} that the angles between the motif elements are all quite similar. This is the shape of the motif in the space if we think of it visually. In the real world example, we have higher dimensional vectors so it is not as easy to visualise.


\subsection{Application}
We apply our method to the Cornell Movie--Dialogs Corpus \cite{danescu2011chameleons} which is a set of conversational movie scripts extracted from 617 movies. We are working with a diverse domain and types of conversations. The language in movie scripts is clean and easy to process, thus allowing us to focus on the pipeline concepts instead of text wrangling. Our use case involves exploring a movie script dataset with particular focus on the comedy genre. 

Using embedded phrases, we construct the graph explained in the methodology section. We found 3232 communities. We drop any communities with less than five members in them. We are left with 1442 communities. We find the centroids of the communities in the original space and reduce the dimensionality by focusing on preserving the cosine similarities between the centroids. We obtain a 0.6 correlation of pairwise cosine distances using UMAP. We obtain 97 sequences after some cleaning mapping to 841 phrase type classes. This occurs because of the phrases that do not map to larger communities which we omit. We extract the longest subsequence of a sequence which has no gaps larger than two consecutive phrases. 

Upon running the algorithm with different initialization, different results are obtained. We detect a global pattern consisting of the following classes: [100, 0, 66, 1, 185]. This has an alignment score of 0.84346919 (with the detected motifs). The underlying classes are interesting and fit a comedy type of script. The community of the detected classes look as follows:

\begin{enumerate}
    \item Huh? [confusion]
    \item Yeah/Yes/Oh/Yep/yes?/yes! [generic type of class allowing for variation - nothing specific]
    \item Tell me/tell me again/that's what you tell me [statement/prompt for further information]
    \item What?/Doing what?/What happened?/The reason?/what! [exclamation of disbelief or question for further information]
    \item You okay? [concern/display of consideration]
\end{enumerate}

 In general, there are slight deviations in individual motifs found across the sequences. The overall tone and message remains quite clear. The motif detected is  of a conversation between characters - some inquiry and an explanation. Individual motifs are expressions of a higher representation of a conversational cue. Since typically it is difficult to separate conversations from emotions, we pick up variations on an emotional theme. An example of what some of these phrases look like can be seen in Table~\ref{tab1}. We see the individual variations in the instance motifs and the overarching patterns.

\begin{table*}[t]

\caption{Table showing an example of motifs detected from comedy movies}\label{tab1}

\begin{tabular}{|p{3,2cm}|p{3,2cm}|p{3,2cm}|p{3,2cm}|p{3,2cm}|}
\hline
\textbf{Motif Position 1} &  \textbf{Motif Position 2} & \textbf{Motif Position 3} & \textbf{Motif Position 4} & \textbf{Motif Position 5}\\
\hline
Well, I hope you've changed. & Yeah, and for yours. I'm sure you've changed. & ...namely a personality. & So how are you? & I'm okay.\\
\hline
The reason? & I see. & Sit down. & What? & Miss Van Cartier.\\
\hline
Huh? & I say we don't want to appear & Oh, that & What do you think you're doing? & Why, I'm assisting you, sir\\
\hline
Do uh... do you understand what I'm saying? & Yes. & Good. Or should I speak slower? & Yes. & Do you follow or should I speak slower?\\
\hline
Are you kidding me? & Seems like a nice place. & It is, if you like idiots... & What do you mean? & Really?\\
\hline
I would have said do. & Aye, but you know him well. & That's what you tell me. & What have you heard? & Is it? Is it really?\\
\hline
\end{tabular}
\end{table*}

\section{Conclusion}
We have shown in our work that it is possible to use motif detection in a different setting in order to extract conversational patterns. We are able to view the results, local alignment scores of the motif and a global pattern. Local alignment scores can assist in selecting the most relevant sequences while the global pattern reveals what the motif abstraction is. In future work, we aim to develop a benchmark and evaluation strategies which are able to better diagnose and quantify results.

\section{Acknowledgements}

The authors want to acknowledge the contribution of ABSA bank which sponsors the Data Science Chair.

%
%
%
%

\end{document}